\documentclass[11pt]{article}

\usepackage[final]{acl}

\usepackage{times}
\usepackage{latexsym}
\usepackage{amsmath}
\usepackage[T1]{fontenc}

\usepackage[utf8]{inputenc}

\usepackage{microtype}

\usepackage{inconsolata}

\usepackage{graphicx}

\usepackage{booktabs}
\usepackage{multirow}
\usepackage[table]{xcolor}
\usepackage{amssymb} 
\usepackage{subcaption}
\title{InfoMerge: Information-aware Token Compression for Efficient Video Large Language Models}

\author{
Xinxin Liu ,
Shiwei Gan\textsuperscript{*} ,
Xiao Liu ,
Yafeng Yin ,
Lei Xie ,
Sanglu Lu \\
State Key Laboratory of Novel Software Technology, Nanjing University \\
Nanjing 210023, China \\
\texttt{231880474@smail.nju.edu.cn,sw@nju.edu.cn, xiaoliu@smail.nju.edu.cn} \\
 \texttt{\{yafeng,lxie,sanglu\}@nju.edu.cn} \\
 \textsuperscript{*}Corresponding author.
}

\begin{document}
\maketitle
\begin{abstract}
Video Large Language Models (Video-LLMs) achieve strong performance in video understanding, but their excessive visual tokens bring substantial computational overhead.
Existing training-free compression methods improve inference efficiency by reducing visual tokens, yet they often rely on local adjacent-frame similarity for temporal redundancy estimation or allocate token budgets mainly according to segment length.
Such designs are sensitive to frame-level noise and fail to capture the non-uniform information distribution of real-world videos.
To address these challenges, we propose \textbf{InfoMerge}, a training-free visual token compression method that improves token utilization through robust redundancy estimation and content-aware budget allocation.
Specifically, we propose the \textbf{Temporal Fingerprint Difference}: a segment-level second-order temporal redundancy estimation strategy, which models the temporal similarity structure of tokens at the same spatial positions within each segment.
We further introduce \textbf{Content-Aware Budget Allocation} (CABA), which dynamically allocates segment-level token budgets based on segment uniqueness and spectral-entropy-based representational richness.
By reducing repeated preservation of redundant static regions and allocating more tokens to informative segments, InfoMerge makes better use of the limited token budget while maintaining strong performance.
Extensive experiments show that InfoMerge achieves strong efficiency--accuracy trade-offs across multiple benchmarks and backbones, with more pronounced advantages under aggressive compression.
On LLaVA-OneVision-7B, InfoMerge retains 98.8\% of the original average performance while reducing 85\% of visual tokens and accelerating the prefill stage by $4.24\times$.
\end{abstract}

\section{Introduction}
\begin{figure}[t]
    \centering

    \begin{subfigure}[t]{0.48\linewidth}
        \centering
        \includegraphics[width=\linewidth]{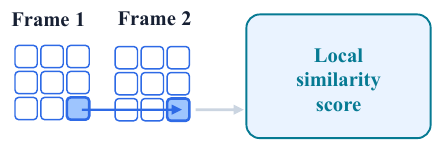}
        \caption{Local similarity estimation.}
        \label{fig:qual_case_left}
    \end{subfigure}
    \hfill
    \begin{subfigure}[t]{0.48\linewidth}
        \centering
        \includegraphics[width=\linewidth]{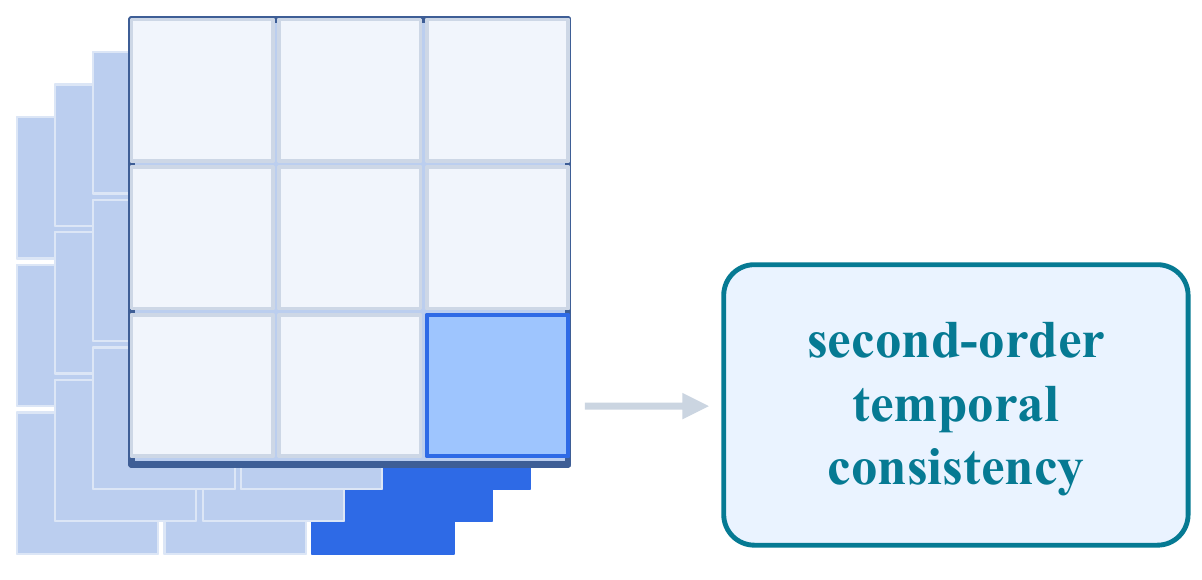}
        \caption{Second-order temporal consistency.}
        \label{fig:qual_case_right}
    \end{subfigure}


    \begin{subfigure}[t]{\linewidth}
        \centering
        \includegraphics[width=\linewidth]{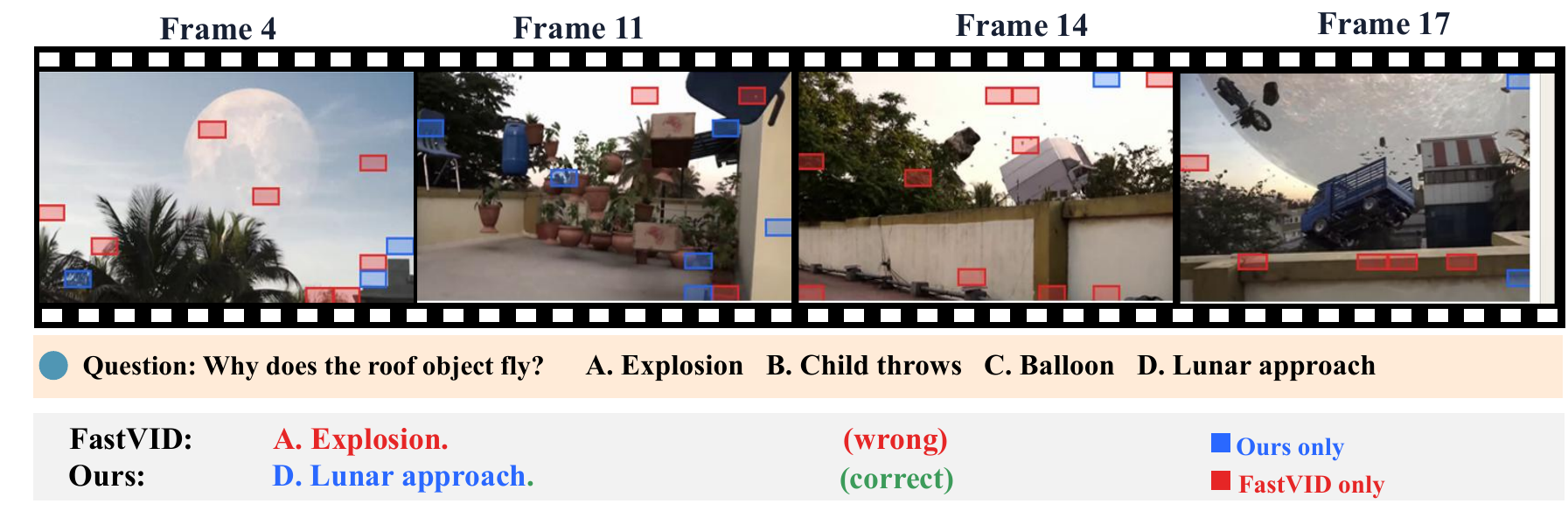}
        \caption{Qualitative comparison between FastVID and ours.}
        \label{fig:qual_case_bottom}
    \end{subfigure}

    \caption{Qualitative analysis of token selection.
    (a) \textbf{Prior methods} rely on local similarity between tokens at the same spatial position in adjacent frames.
    (b) We further model second-order temporal consistency across frames.
    (c) Compared with FastVID, our method retains more informative tokens related to the flying roof object, leading to the correct answer.}
    \label{fig:qualitative_case}
\end{figure}

Video-LLMs have achieved remarkable progress in complex video understanding tasks~\cite{li2024llava-ov,bai2025qwen3,li2025videochat,cheng2024videollama,wang2024qwen2,zhang2024llava-video}.
Despite their strong capabilities, the high computational and memory costs induced by dense video tokens make token compression essential for efficient Video LLM inference.
Existing training-free methods can be broadly categorized into spatial pruning methods~\cite{yang2025visionzip,chen2024image}, temporal redundancy reduction methods~\cite{tao2025dycoke}, staged spatio-temporal compression methods~\cite{shen2026fastvid,shao2025holitom}, and unified spatio-temporal approaches~\cite{du2026unified,zhang2026d2pruner}. Despite their progress, most existing methods rely on a common underlying assumption: \textit{local pairwise similarity is sufficient to characterize redundancy}. In practice, however, this assumption becomes fragile under real-world video noise, motion, and viewpoint variations, leading to unstable redundancy estimation and suboptimal token allocation.

Moreover, segment-based methods such as FastVID~\cite{shen2026fastvid} allocate token budgets mainly according to segment length, ignoring the intrinsic information content of each segment.
Based on these observations, we reformulate video token compression as a problem of second-order temporal redundancy estimation and information-aware budget allocation, aiming to reduce repeated preservation of static tokens and assign more budget to semantically distinctive and representation-rich segments.

To this end, we propose \textbf{InfoMerge}, a training-free video token compression method from a segment-structured redundancy perspective.
Unlike previous methods that estimate temporal redundancy using first-order adjacent-frame similarity at the same spatial position, InfoMerge introduces \textbf{Temporal Fingerprint Difference (TFD)} to evaluate temporal redundancy from a second-order perspective by modeling pairwise similarity structures within each segment.
We further propose \textbf{Content-Aware Budget Allocation (CABA)}, which jointly considers segment-level uniqueness and internal representation complexity to adaptively allocate token budgets according to semantic distinctiveness and spectral information density.

Finally, we integrate TFD and CABA into a spatio-temporal compression pipeline that performs redundancy-aware token merging in a training-free and plug-and-play manner. The proposed method can be seamlessly integrated into existing VLLMs.
Figure~\ref{fig:qualitative_case} qualitatively shows that InfoMerge retains more informative visual evidence than FastVID under the same retention ratio, which helps the model produce the correct answer.
Extensive experiments demonstrate that InfoMerge achieves a superior efficiency–accuracy trade-off.
Our contributions are summarized as follows:

\begin{itemize}
    \item We propose \textbf{Temporal Fingerprint Difference}, which estimates temporal redundancy from a structured second-order perspective and provides a robust static-token prior for more efficient token selection.

    \item We introduce \textbf{Content-Aware Budget Allocation}, which adaptively allocates token budgets according to segment-level uniqueness and representational richness, assigning more tokens to informative segments.

    \item  We propose InfoMerge, a training-free and plug-and-play method that can be seamlessly integrated into existing Video-LLMs without fine-tuning. InfoMerge achieves strong efficiency–accuracy trade-offs, retaining most of the original model performance even under extremely low token retention ratios.
    
\end{itemize}

\section{Related Work}
\paragraph{Video Large Language Models.}

Recent Video-LLMs~\cite{li2025videochat,cheng2024videollama,li2024llava-next,li2024llava-ov,zhang2024llava-video,bai2025qwen3,wang2025internvl3}
have achieved strong performance in video understanding by extending image-based VLMs to video inputs.
However, these models still pass thousands of visual tokens to the downstream LLM, leading to substantial computational overhead due to the quadratic complexity of attention~\cite{vaswani2017attention}.
Although some recent works~\cite{lin2024vila,liu2025nvila} improve token efficiency through model-level optimization, they typically require additional training or fine-tuning.
Therefore, efficient training-free token compression remains important for practical VLLM inference.

\paragraph{Visual Token Compression.}

\begin{figure*}[t]
    \centering
    \includegraphics[width=0.95\textwidth]{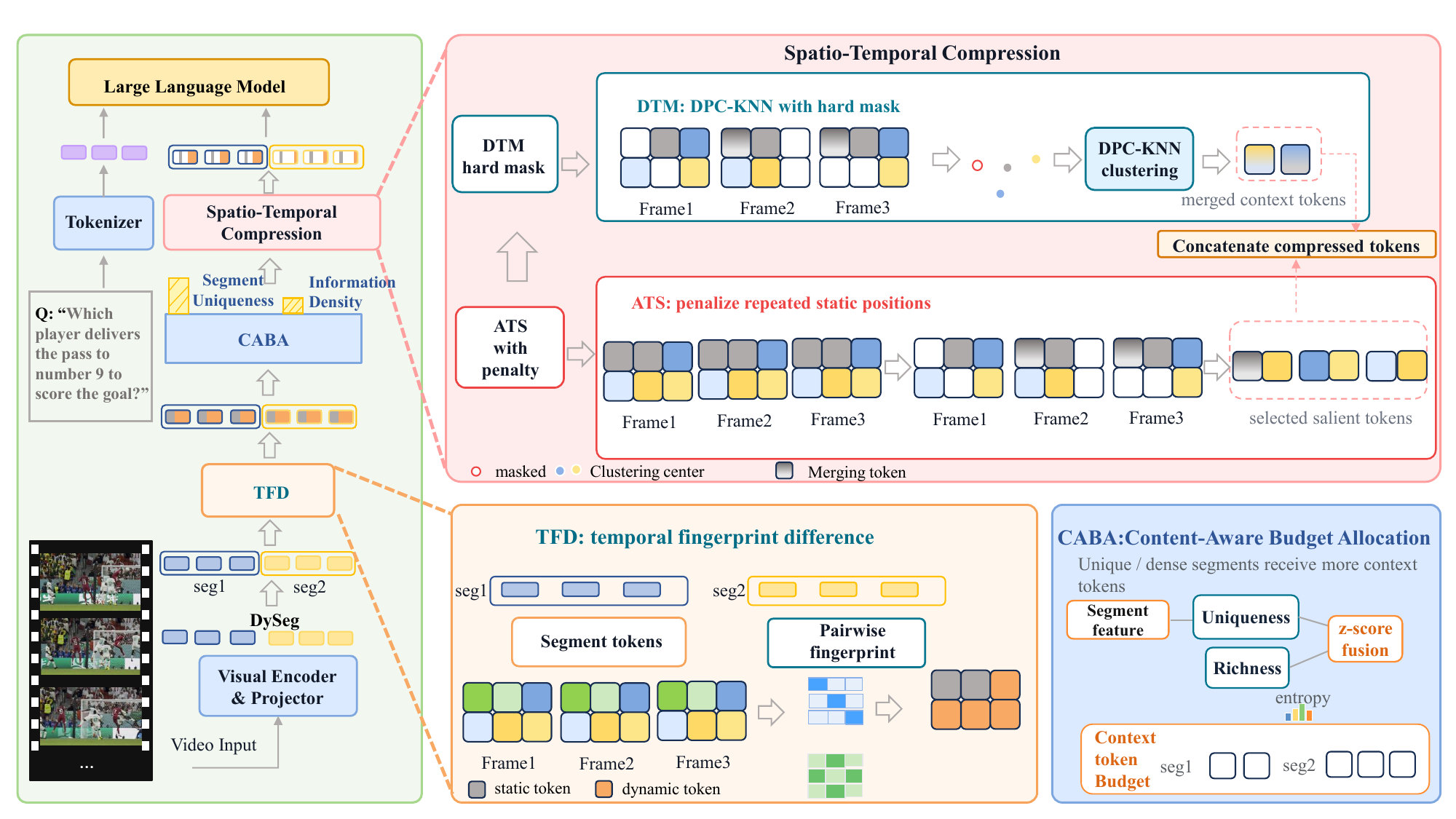}
    \caption{Overview of the proposed InfoMerge method.}
    \label{fig:method}
\end{figure*}

Token compression is an effective way to reduce token redundancy in ViTs and LLMs. Training-free visual token compression aims to reduce the inference cost of VLMs and Video-LLMs without additional model training. Existing methods mainly rely on token pruning or merging. Spatial methods such as ToMe~\cite{bolya2022tome}, VisionZip~\cite{yang2025visionzip}, and FastV~\cite{chen2024image} reduce visual tokens according to token similarity or attention scores. Video-oriented methods further exploit temporal redundancy: DyCoke~\cite{tao2025dycoke}, TempMe~\cite{shen2025tempme}, and PruneVID~\cite{huang2025prunevid} merge similar tokens across frames, while FastVID~\cite{shen2026fastvid} performs segmentation-based token selection and merging. HoliTom~\cite{shao2025holitom} further considers global redundancy for temporal token organization.
However, two limitations remain. First, temporal redundancy is often characterized by first-order local cues, such as adjacent-frame similarity or attention scores. These cues can be unstable under camera motion, illumination changes, or local perturbations, and they do not explicitly model whether a region is temporally static or dynamically informative. Second, segment-based methods usually allocate token budgets mainly according to segment length, which cannot capture the non-uniform information density of real videos. In contrast, our method uses Temporal Fingerprint Difference to model second-order temporal consistency within each segment, and Content-Aware Budget Allocation to allocate more tokens to semantically unique and representation-rich segments.

\section{Method}
\label{sec:method}
\paragraph{Overview.}
\label{sec:method:overview}
As illustrated in Fig.~\ref{fig:method}, InfoMerge consists of four components:
dynamic temporal segmentation, Temporal Fingerprint Difference (TFD),
Content-Aware Budget Allocation (CABA), and spatio-temporal compression.
Given sampled video tokens, we first divide frames into temporally coherent segments.
TFD then identifies temporally redundant spatial positions within each segment,
while CABA adaptively allocates token budgets according to segment information content.
Finally, we integrate both the TFD prior and CABA-guided segment budgets into spatio-temporal compression to suppress repeated preservation of static regions.

\subsection{Dynamic Temporal Segmentation}
\label{sec:method:dyseg}



Given a video consisting of $L$ frames, where each frame contains $N$ visual tokens with hidden dimension $D$, we first apply Dynamic Temporal Segmentation (DySeg) to divide the sampled video frames into temporally coherent segments.
DySeg estimates transition strength using the cosine similarity between adjacent frame-level global features and selects low-similarity transitions as segment boundaries.
This produces a sequence of ordered segments $\{\mathcal{V}_k\}_{k=1}^{K}$.
For each segment $\mathcal{V}_k$, we use local frame indices $l\in\{1,\dots,|\mathcal{V}_k|\}$, where $|\mathcal{V}_k|$ denotes the segment length.
These segments serve as the basic units for our subsequent TFD, CABA, and Spatio-Temporal Compression.

\subsection{Temporal Fingerprint Difference}
\begin{figure}[t]
    \centering
    \includegraphics[
        width=\linewidth,
        trim=5 55 5 65,
        clip
    ]{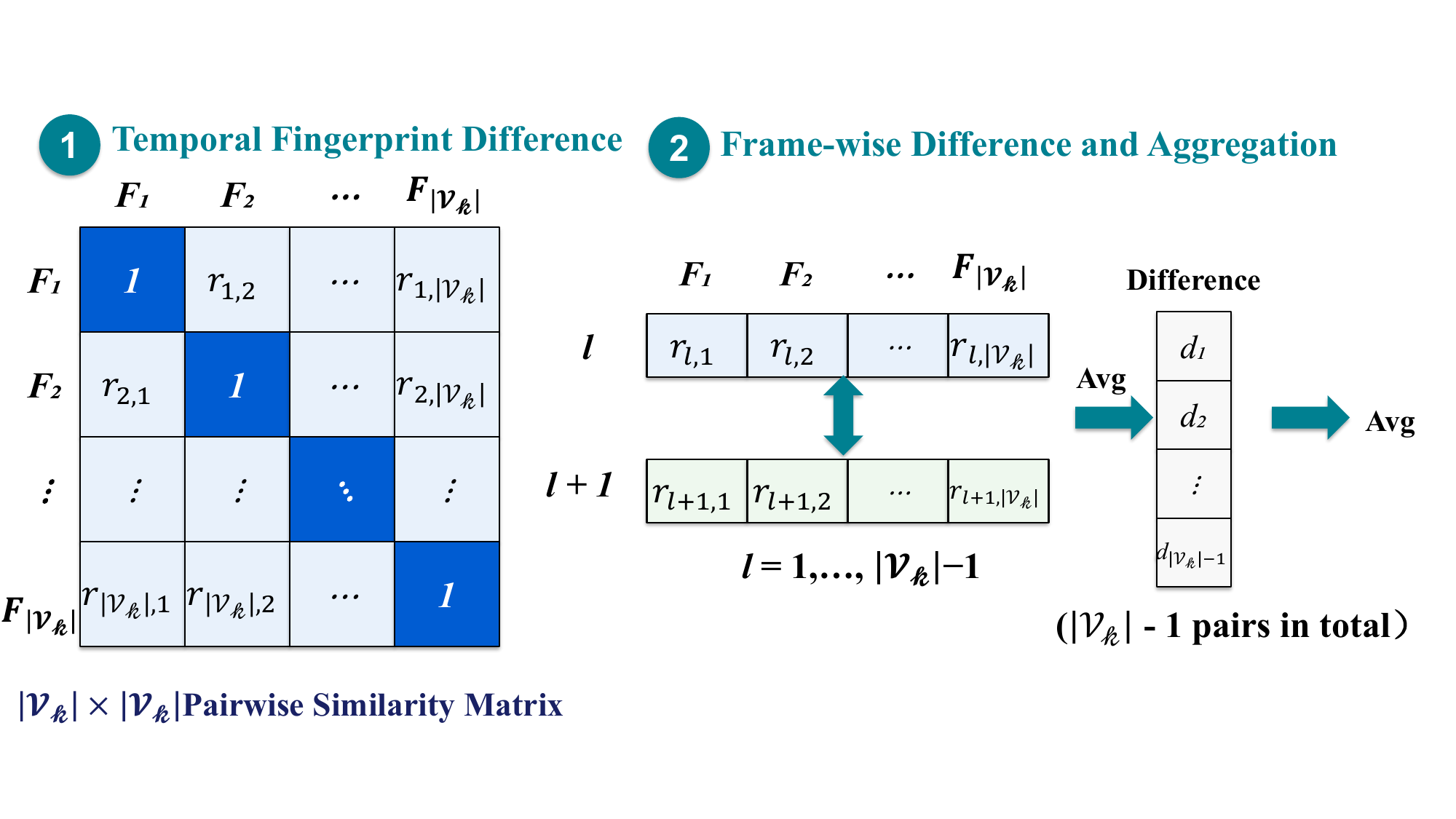}
    \caption{Illustration of the proposed TFD. For each spatial position, we construct an intra-segment temporal similarity matrix and compute the differences between adjacent temporal fingerprints to estimate temporal dynamics. } 
    \label{fig:madd}
    \vspace{-3mm}
\end{figure}
Existing methods~\cite{tao2025dycoke,shao2025holitom} estimate temporal redundancy mainly through local similarities between adjacent frames. However, such short-range measurements are sensitive to frame-level noise and often fail to reliably distinguish temporally static tokens from dynamically informative ones. We argue that truly redundant tokens should exhibit globally consistent temporal patterns within a segment, rather than only local frame-to-frame similarity. 

To this end, we propose a \textbf{Temporal Fingerprint Difference (TFD)} module, which models segment-level temporal consistency for robust redundancy estimation, as illustrated in Fig.~\ref{fig:madd}. 
The design is conceptually inspired by MADD~\cite{sarkar2019perfect}, which characterizes high-dimensional samples according to their relations to the data cloud.
TFD constructs a segment-level second-order temporal fingerprint for each spatial token position. 

Specifically, given video features $\mathbf{H}\in\mathbb{R}^{L\times N\times D}$, let $\mathbf{H}^{(k)}_{l,p}\in\mathbb{R}^{D}$ denote the token at spatial position $p$ in frame $l$ of the $k$-th segment $\mathcal{V}_k$.
For each spatial position $p$, we first normalize token features across frames and then construct a temporal similarity fingerprint matrix:
\begin{equation}
\small
\mathbf{F}^{(k)}_p[l,l']
=
\mathrm{Cos}
\left(
\frac{\mathbf{H}^{(k)}_{l,p}}{\|\mathbf{H}^{(k)}_{l,p}\|_2},
\frac{\mathbf{H}^{(k)}_{l',p}}{\|\mathbf{H}^{(k)}_{l',p}\|_2}
\right)
\end{equation}
where $l,l' \in\{1,\dots,|\mathcal{V}_k|\}$, and $\mathrm{Cos}(\cdot,\cdot)$ denotes the cosine similarity function and
$\mathbf{F}^{(k)}_p\in\mathbb{R}^{|\mathcal{V}_k|\times |\mathcal{V}_k|}$ characterizes the pairwise temporal similarity structure of spatial position $p$ throughout the entire segment. 
Intuitively, temporally static regions tend to maintain stable similarity structures across time, while dynamic regions exhibit larger temporal variations. We measure temporal dynamics by computing the average absolute difference between adjacent temporal fingerprints, corresponding to adjacent rows of the fingerprint matrix:
\begin{equation} 
\resizebox{0.9\linewidth}{!}{$
\mathrm{TFD}^{(k)}(p)
=
\frac{
\sum_{l=1}^{|\mathcal{V}_k|-1}
\sum_{l'=1}^{|\mathcal{V}_k|}
\left|
\mathbf{F}^{(k)}_p[l+1,l']
-
\mathbf{F}^{(k)}_p[l,l']
\right|} {(|\mathcal{V}_k|-1)|\mathcal{V}_k|}
$}
\end{equation}
A smaller TFD value indicates stronger temporal consistency and thus higher redundancy.
Accordingly, for each segment, we select the $\lfloor \rho N \rfloor$ spatial positions with the smallest TFD scores as static candidates $\mathcal{S}^{(k)}$.

\subsection{Content-Aware Budget Allocation}
\label{sec:method:budget}

Video information is often unevenly distributed across time: long segments may contain mostly static or repetitive content, while short segments can include key actions or scene changes.
Therefore, allocating token budgets only according to segment length may under-preserve short but informative segments.
To address this issue, we propose \textbf{Content-Aware Budget Allocation (CABA)}, which estimates segment-level information content from segment uniqueness and representational richness.
\paragraph{Segment Uniqueness.}
We compute the segment-level representation of the $k$-th segment $\mathcal{V}_k$ as:
\begin{equation}
\bar{\mathbf{x}}_k
=
\frac{1}{|\mathcal{V}_k|N}
\sum_{l=1}^{|\mathcal{V}_k|}
\sum_{p=1}^{N}
\mathbf{H}^{(k)}_{l,p},
\end{equation}
Further, we compute the global video representation $\bar{\mathbf{x}}_{\mathrm{glob}}$:
\begin{equation}
\bar{\mathbf{x}}_{\mathrm{glob}}=\frac{\sum_{k=1}^{K}|\mathcal{V}_k|\bar{\mathbf{x}}_k}{\sum_{k=1}^{K}|\mathcal{V}_k|}.
\end{equation}
We define the segment uniqueness score as the cosine distance between the segment representation and the global video representation:
\begin{equation}
u_k=1-\mathrm{Cos}(\bar{\mathbf{x}}_k,\bar{\mathbf{x}}_{\mathrm{glob}}).
\end{equation}
A larger $u_k$ indicates that the segment is more semantically distinctive from the overall video.

\paragraph{Representational Richness.}
\begin{figure}[t]
    \centering
    \includegraphics[width=\linewidth]{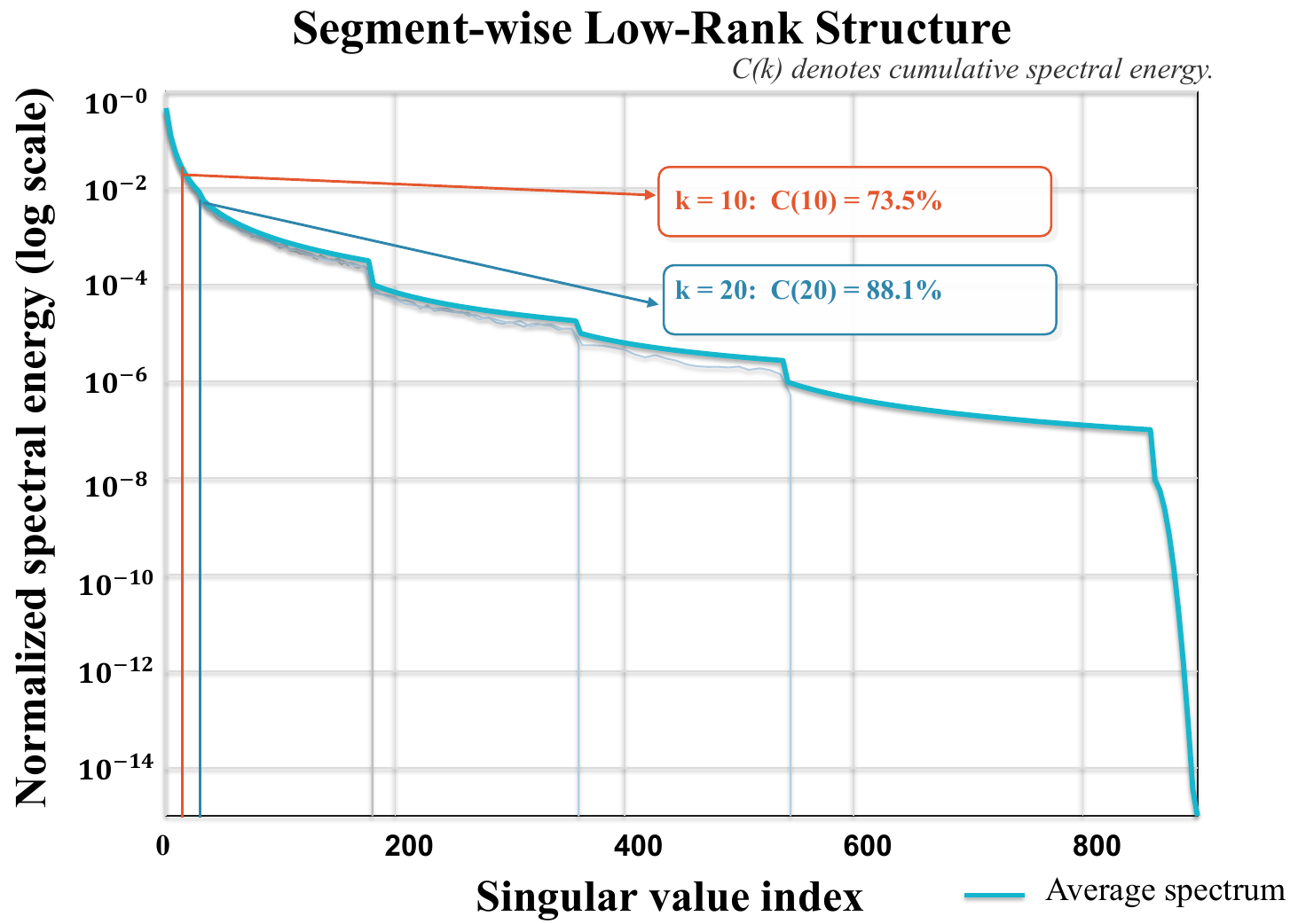}
    \caption{
\textbf{Analysis of segment-wise low-rank structures in video token representations.}
The rapidly decaying singular-value spectra indicate strong redundancy within the segment.
}
\label{fig:erank}
\end{figure}

For the $k$-th segment  $\mathcal{V}_k$, we flatten its visual tokens into
$\mathbf{X}_k \in \mathbb{R}^{|\mathcal{V}_k|N \times D}$ and compute its singular values $\{\sigma_j\}_{j=1}^{R}$, where $R=\min(|\mathcal{V}_k|N,D)$.
As shown in Fig.~\ref{fig:erank}, video token representations exhibit a clear low-rank structure within a segment.The top-20 singular directions already explain 88.1\% of the spectral energy.
This motivates the use of spectral entropy to estimate segment-level representational richness.
We estimate representational richness using normalized spectral entropy:
\begin{equation}
\pi_j
=
\frac{\sigma_j^2}
{\sum_{\ell=1}^{R}\sigma_\ell^2}
\end{equation}
We then compute the normalized spectral entropy:
\begin{equation}
e_k
=
\frac{
-\sum_{j=1}^{R}
\pi_j \log \pi_j
}{
\log R
}.
\end{equation}
A larger $e_k$ indicates that the segment spans more diverse principal directions and is therefore less likely to be represented by a few redundant patterns.

\paragraph{Budget Allocation.} We fuse segment uniqueness and representational richness after z-score normalization as $m_k=\alpha z(u_k)+\beta z(e_k)$.
where $\alpha$ and $\beta$ control the contributions of the two factors.
We then use a sigmoid function to map the fused score to a smooth budget modulation coefficient and multiply it by the segment length:
\begin{equation}
w_k = \operatorname{sigmoid}(\tau m_k)\cdot |\mathcal{V}_k|,
\end{equation}
where $\tau$ is the temperature parameter that controls the sharpness of budget redistribution.
Finally, the context-token budget for the $k$-th segment is computed as
\begin{equation}
B_k
=
\max
\left(
1,
\operatorname{round}
\left(
\frac{L T_c w_k}
{\sum_{j=1}^{K} w_j}
\right)
\right),
\end{equation}
where  $T_c$ denotes the average context-token budget per frame.
In this way, CABA allocates more tokens to semantically unique and representation-rich segments.
\begin{table*}[t]
\centering
\small
\setlength{\tabcolsep}{3pt}
\renewcommand{\arraystretch}{0.95}
\caption{\textbf{Comparison on LLaVA-OneVision-7B.}
The A\%/B\% retention ratio indicates that A\% of the LLM input tokens are retained, and subsequently compressed to B\% during the LLM forward pass.
The Avg. Score is computed over VideoMME Overall, MLVU, and LongVideoBench.
\textbf{Best} results are in bold, \underline{second best} underlined.}
\label{tab:sota_ov7b}

\resizebox{\textwidth}{!}{%
\begin{tabular}{l c c c c c c c c c c c}
\toprule
\multirow{2}{*}{Method}
& \multirow{2}{*}{\shortstack{FLOPs\\(T)\,$\downarrow$}}
& \multirow{2}{*}{\shortstack{FLOPs\\Ratio}}
& \multirow{2}{*}{\shortstack{Retention\\Ratio}}
& \multicolumn{4}{c}{VideoMME\,$\uparrow$}
& \multirow{2}{*}{MLVU\,$\uparrow$}
& \multirow{2}{*}{\shortstack{Long\\VideoBench\,$\uparrow$}}
& \multicolumn{2}{c}{Avg.\,$\uparrow$} \\
\cmidrule(lr){5-8} \cmidrule(lr){11-12}
& & & & Short & Medium & Long & Overall & & & Score & \% \\
\midrule
\rowcolor{gray!12}
LLaVA-OV-7B~\cite{li2024llava-ov}
& 48.8 & 100.0\% & 100\%
& 70.2 & 56.8 & 48.8 & 58.6
& 64.9 & 56.4
& 59.97 & 100.0 \\
\midrule
FastV~\cite{chen2024image}
& 9.4 & 19.2\% & 100\%/15\%
& 65.6 & 55.1 & 46.9 & 55.9
& 60.2 & 53.2
& 56.43 & 94.1 \\
VisionZip~\cite{yang2025visionzip}
& 6.3 & 12.9\% & 15\%
& 67.4 & \textbf{57.6} & \textbf{48.1} & 57.7
& 60.0 & 55.8
& 57.83 & 96.4 \\
MMG-Vid~\cite{ma2026mmg}
& 6.3 & 12.9\% & 15\%
& -- & -- & -- & \underline{57.9}
& 61.6 & 55.9
& 58.47 & 97.5 \\
FastVID~\cite{shen2026fastvid}
& 6.3 & 12.9\% & 15\%
& \underline{69.3} & 56.4 & 47.6 & 57.8
& \underline{62.9} & \underline{56.2}
& \underline{58.97} & \underline{98.3} \\
\textbf{Ours}
& 6.3 & 12.9\% & 15\%
& \textbf{69.6} & \underline{56.8} & \textbf{48.1} & \textbf{58.2}
& \textbf{63.1} & \textbf{56.4}
& \textbf{59.23} & \textbf{98.8} \\
\midrule
FastV~\cite{chen2024image}
& 7.4 & 15.1\% & 100\%/10\%
& 61.7 & 54.1 & 47.1 & 54.3
& 58.6 & 52.1
& 55.00 & 91.7 \\
VisionZip~\cite{yang2025visionzip}
& 4.2 & 8.5\% & 10\%
& 63.6 & \textbf{56.0} & \textbf{48.2} & 55.9
& 59.7 & 54.5
& 56.70 & 94.5 \\
FastVID~\cite{shen2026fastvid}
& 4.2 & 8.5\% & 10\%
& \underline{67.2} & 55.7 & \underline{47.6} & \underline{56.8}
& \textbf{62.1} & \textbf{55.7}
& \textbf{58.20} & \textbf{97.1} \\
\textbf{Ours}
& 4.2 & 8.5\% & 10\%
& \textbf{68.9} & \underline{55.9} & \textbf{48.2} & \textbf{57.7}
& \underline{61.7} & \underline{55.2}
& \textbf{58.20} & \textbf{97.1} \\
\midrule
FastV~\cite{chen2024image}
& 5.4 & 11.1\% & 100\%/5\%
& 56.3 & 51.7 & 44.1 & 50.7
& 56.1 & 48.5
& 51.77 & 86.3 \\
VisionZip~\cite{yang2025visionzip}
& 2.1 & 4.2\% & 5\%
& 59.8 & 51.7 & \textbf{45.9} & 52.4
& 58.5 & 48.4
& 53.10 & 88.5 \\
FastVID~\cite{shen2026fastvid}
& 2.1 & 4.2\% & 5\%
& \underline{63.3} & \underline{51.8} & 45.6 & \underline{53.6}
& \underline{58.8} & \underline{51.4}
& \underline{54.60} & \underline{91.0} \\
\textbf{Ours}
& 2.1 & 4.2\% & 5\%
& \textbf{65.2} & \textbf{53.0} & \underline{45.7} & \textbf{54.6}
& \textbf{59.5} & \textbf{52.1}
& \textbf{55.40} & \textbf{92.4} \\
\bottomrule
\end{tabular}%
}
\end{table*}

\subsection{Spatio-Temporal Compression}
\label{sec:method:compression}

We follow the Attention-Based Token Selection (ATS)--Density-Based Token Merging (DTM) compression pipeline, where the TFD-estimated static prior guides redundancy-aware token selection and merging, while the CABA-guided segment budgets adaptively control the context-token allocation across segments.

\paragraph{Attention-Based Token Selection with Penalty.}

We use the attention from the [CLS] token as a saliency estimate. For SigLIP-based Video-LLMs, we follow FastVID~\cite{shen2026fastvid} to obtain these attention scores with the pretrained SigLIP head.
However, high-attention static regions may be repeatedly selected across frames.
To alleviate this issue, we introduce a redundancy-sensitive suppression mechanism guided by the static set of candidates $\mathcal{S}^{(k)}$. 
For frame $l$ in segment $\mathcal{V}_k$, let $\mathbf{A}_l\in\mathbb{R}^{N}$ denote flattened attention scores.
We maintain a historical static set $\mathcal{P}^{(k)}_{l-1}$, which records the static token positions selected by ATS in the previous $l-1$ frames of the current segment.
We only penalize positions that are both static candidates and have already been selected:
\begin{equation}
\begin{aligned}
\mathbf{A}_{l,p}
&\leftarrow
\mathbf{A}_{l,p}
-
\lambda \cdot \operatorname{std}(\mathbf{A}_l), \\
&\hspace{1.5em}
p\in
\mathcal{S}^{(k)}
\cap
\mathcal{P}^{(k)}_{l-1}.
\end{aligned}
\end{equation}
where $\lambda$ controls the redundancy suppression strength.
ATS then selects the top positions $T_s$ according to the updated attention scores, forming the salient-token set $\mathcal{T}^{\mathrm{ATS,(k)}}_l$, where $T_s$ denotes the salient-token budget per-frame.
If a selected position belongs to the static candidate set, it is added to the historical static set:
\begin{equation}
\mathcal{P}^{(k)}_{l}
=
\mathcal{P}^{(k)}_{l-1}
\cup
\left(
\mathcal{T}^{\mathrm{ATS,(k)}}_l
\cap
\mathcal{S}^{(k)}
\right).
\end{equation}
For these selected static positions, we further replace their token representations in the current and subsequent frames of the segment with a segment-level merged representation at the same spatial position. Details are provided in Appendix ~\ref{app:ats}.

\paragraph{Hard-Masked Density-Based Token Merging.}


For each segment, we sample anchor frames at a fixed interval $p$ and evenly distribute the segment-level context-token budget across these anchor frames.
Within each anchor frame, for token position $p$, DTM computes a density-peak score $q^{(k)}_{a,p}$, where $a$ denotes the index of the anchor frame in segment $\mathcal{V}_k$.
High-scoring tokens are selected as context anchors.
However, without additional constraints, multiple anchor frames may repeatedly select the same static regions as cluster centers, leading to redundant context preservation.
To improve contextual diversity, we introduce a hard mask.
Let $\mathcal{Q}^{(k)}_{a-1}$ denote the set of static positions already selected by the previous $a-1$ anchor frames within $\mathcal{V}_k$.
For the current anchor frame $a$, if position $p$ belongs to $\mathcal{Q}^{(k)}_{a-1}$, its density-peak score is directly suppressed:
\begin{equation}
q^{(k)}_{a,p}
\leftarrow
-\infty,
\quad
p\in\mathcal{Q}^{(k)}_{a-1}.
\end{equation}
The remaining high-scoring positions are then selected as anchor tokens according to the allocated context-token budget.
After anchor selection, non-anchor tokens are assigned to their nearest anchor centers and aggregated into the corresponding anchor tokens  through weighted merging.
The salient tokens preserved by ATS and the contextual tokens generated by DTM are then concatenated in temporal order to form the compressed video-token sequence.

\section{Experiments}
\label{sec:exp}

\begin{table*}[t]
\centering
\small
\setlength{\tabcolsep}{4.5pt}
\caption{\textbf{Comparison on LLaVA-Video-7B-Qwen2.}
We report results on MVBench and VideoMME.
FLOPs are measured under the default 64-frame setting.
``\%'' denotes the percentage of the Vanilla average score retained.}
\label{tab:llava_video_sota}
\resizebox{\textwidth}{!}{
\begin{tabular}{l c c c c c c c c c c}
\toprule
\multirow{2}{*}{Method} 
& \multirow{2}{*}{\shortstack{Retention\\Ratio $R$}} 
& \multirow{2}{*}{\shortstack{FLOPs\\(T)$\downarrow$}} 
& \multirow{2}{*}{\shortstack{FLOPs\\Ratio}} 
& \multirow{2}{*}{MVBench} 
& \multicolumn{4}{c}{VideoMME} 
& \multirow{2}{*}{\shortstack{Avg.\\Score}} 
& \multirow{2}{*}{\%} \\
\cmidrule(lr){6-9}
& & & & & Overall & Short & Medium & Long & & \\
\midrule
\rowcolor{gray!12}
LLaVA-Video-7B
& 100\% & 94.1 & 100.0\% 
& 60.4 & 64.1 & 77.0 & 62.3 & 53.1 
& 62.3 & 100.0 \\

\midrule
FastV~\cite{chen2024image}
& 15\% & 16.9 & 18.0\%
& 50.8 
& 54.0 & 60.9 & 54.4 & 46.7 
& 52.4 & 84.2 \\

VisionZip~\cite{yang2025visionzip}
& 15\% & 11.0 & 11.7\%
& 55.2 
& 60.3 & 70.3 & 58.8 & \underline{51.7}
& 57.8 & 92.8 \\

MMG-Vid~\cite{ma2026mmg}
& 15\% & 11.0 & 11.7\%
& 56.1 
& 61.1 & 72.3 & 60.1 & 50.8
& 58.6 & 94.1 \\

FastVID~\cite{shen2026fastvid}
& 15\% & 11.0 & 11.7\%
& \textbf{60.5}
& \underline{62.1} & \textbf{73.8} & \underline{61.0} & \underline{51.7}
& \underline{61.3} & \underline{98.5} \\

\textbf{Ours} 
& 15\% & 11.0 & 11.7\%
& \underline{59.9} 
& \textbf{62.8} & \underline{73.4} & \textbf{63.0} & \textbf{51.9}
& \textbf{61.4} & \textbf{98.6} \\

\midrule
FastV~\cite{chen2024image}
& 10\% & 13.1 & 13.9\%
& 43.2 
& 49.6 & 54.0 & 50.3 & 44.6
& 46.4 & 74.5 \\

VisionZip~\cite{yang2025visionzip}
& 10\% & 6.9 & 7.3\%
& 53.8 
& 58.7 & 67.4 & 57.7 & \textbf{51.1}
& 56.3 & 90.4 \\

MMG-Vid~\cite{ma2026mmg}
& 10\% & 6.9 & 7.3\%
& 54.9 
& 59.4 & 71.0 & 57.9 & 49.2
& 57.2 & 91.8 \\

FastVID~\cite{shen2026fastvid}
& 10\% & 6.9 & 7.3\%
& \underline{58.5} 
& \underline{60.2} & \underline{72.0} & \underline{58.1} & 50.4
& \underline{59.4} & \underline{95.3} \\

\textbf{Ours} 
& 10\% & 6.9 & 7.3\%
& \textbf{60.0} 
& \textbf{60.9} & \textbf{72.6} & \textbf{59.1} & \underline{51.0}
& \textbf{60.5} & \textbf{97.1} \\

\bottomrule
\end{tabular}
}
\end{table*}

\subsection{Experimental Settings}
\label{sec:exp_setup}

\paragraph{Benchmarks and Baselines.}
We evaluate InfoMerge with LMMS-Eval~\cite{li2024xinrun,zhang2025lmms} on three widely used video understanding benchmarks: VideoMME~\cite{fu2025videomme}, MLVU~\cite{zhou2025mlvu}, and LongVideoBench~\cite{wu2024longvideobench}.
LLaVA-OneVision-7B~\cite{li2024llava-ov} is used as the main evaluation backbone, and the original uncompressed model serves as the upper-bound baseline. To evaluate cross-backbone transferability, we further apply InfoMerge to LLaVA-Video-7B~\cite{zhang2024llava-video} and report results on VideoMME and MVBench.
We compare InfoMerge with representative training-free visual token compression methods, including FastV~\cite{chen2024image}, VisionZip~\cite{yang2025visionzip}, and FastVID~\cite{shen2026fastvid}, MMG-Vid~\cite{ma2026mmg}. 
\paragraph{Implementation Details.}
We follow the default setting and sample 32 frames per video for LLaVA-OneVision-7B~\cite{li2024llava-ov}.
For LLaVA-Video-7B~\cite{zhang2024llava-video}, we sample 64 frames by default.
All experiments are conducted on one NVIDIA V100 GPU.
We evaluate different compression strengths by setting the visual token retention ratios to 5\%, 10\%, and 15\%.
Unless otherwise specified, all ablation studies are conducted on LLaVA-OneVision-7B using the same frame sampling strategy, inference parameters, and evaluation protocol. Detailed hyperparameter settings are provided in Appendix~\ref{app:hy}.
\subsection{Comparisons}
\label{sec:main_results}
\paragraph{Performance on LLaVA-OneVision-7B.}
Table~\ref{tab:sota_ov7b} compares InfoMerge with representative training-free token compression methods on LLaVA-OneVision-7B.
Under the same retention ratio and FLOPs, InfoMerge improves over FastVID at 15\% and 5\% retention ratios, and achieves comparable overall performance at 10\% retention.
At the 15\% retention ratio, InfoMerge retains 98.8\% of the uncompressed model performance with only 12.9\% FLOPs.
Compared with the recent MMG-Vid~\cite{ma2026mmg}, InfoMerge improves the retained performance from 97.5\% to 98.8\%.
On VideoMME, InfoMerge also achieves the highest overall score of 58.2, outperforming both FastVID~\cite{shen2026fastvid} and MMG-Vid~\cite{ma2026mmg}.
At the 10\% retention ratio, InfoMerge shows clear gains on VideoMME.
It improves the VideoMME overall score from 56.8 to 57.7 over FastVID~\cite{shen2026fastvid}, yielding a 0.9-point improvement under the same FLOPs.
InfoMerge improves the average score over FastV by 3.20 points.
At the more aggressive 5\% retention ratio, InfoMerge still maintains strong performance.
Specifically, with 95\% of visual tokens compressed, InfoMerge retains 92.4\% of the original average performance, whereas the best-performing baseline FastVID retains 91.0\%.
Compared with the baseline methods, InfoMerge achieves consistently competitive overall performance across different retention ratios, with more pronounced advantages under extremely low token retention ratios.

\paragraph{Cross-backbone Transfer.} 

Table~\ref{tab:llava_video_sota} reports results on LLaVA-Video-7B.
InfoMerge retains 97.1\% of the original average performance with only 10\% of visual tokens preserved, outperforming FastVID, which retains 95.3\% under the same setting.
This result suggests that InfoMerge can transfer to another Video-LLM backbone.

\paragraph{Efficiency Comparison.}
\begin{table}[t]
\centering
\small
\setlength{\tabcolsep}{3pt}
\renewcommand{\arraystretch}{0.9}
\caption{
LLM-side latency comparison.
Prefill time measures LLM decoder latency until the first generated token.
Generate time measures LLM decoder latency for the full response, including prefill and subsequent decoding.
}
\label{tab:table2-phase-latency}
\resizebox{\linewidth}{!}{
\begin{tabular}{lccc}
\toprule
\multirow{2}{*}{\textbf{Method}} 
& \multicolumn{2}{c}{\textbf{Time}} 
& \multirow{2}{*}{\textbf{Acc. (\%)}} \\
\cmidrule(lr){2-3}
& \textbf{Prefill (ms)} & \textbf{Generate (ms)} & \\
\midrule
LLaVA-OV-7B      & 1505.7 (1.00$\times$)  & 1538.5 (1.00$\times$)  & 100.0 \\
Ours $r{=}15\%$ & 355.4 (4.24$\times$)   & 389.2 (3.95$\times$)   & 98.8 \\
Ours $r{=}10\%$ & 225.6 (6.67$\times$)   & 257.6 (5.97$\times$)   & 97.1 \\
Ours $r{=}5\%$  & 121.1 (12.43$\times$)  & 150.3 (10.23$\times$)  & 92.4 \\
\bottomrule
\end{tabular}}
\end{table}

\begin{table}[t]
\centering
\small
\setlength{\tabcolsep}{5pt}
\renewcommand{\arraystretch}{0.95}
\caption{Ablation study on different components.}
\label{tab:component_ablation}
\begin{tabular}{ccc|ccc}
\toprule
\multicolumn{3}{c|}{\textbf{Components}} 
& \multicolumn{3}{c}{\textbf{VideoMME Acc.}} \\
\cmidrule(lr){1-3} \cmidrule(lr){4-6}
\textbf{TFD} & \textbf{Unique} & \textbf{Richness} 
& \textbf{$r{=}5\%$} & \textbf{$r{=}10\%$} & \textbf{$r{=}15\%$} \\
\midrule
           &            &            & 53.6 & 56.8 & 57.8 \\
\checkmark &            &            & 54.2 & 57.5 & 58.1 \\
\checkmark & \checkmark &            & 54.5 & 57.6 & 58.1 \\
\checkmark &            & \checkmark & 54.3 & 56.9 & \textbf{58.2} \\
\checkmark & \checkmark & \checkmark & \textbf{54.6} & \textbf{57.7} & \textbf{58.2} \\
\bottomrule
\end{tabular}
\end{table}
Beyond performance,
Table~\ref{tab:table2-phase-latency} reports the LLM-side latency after visual features are obtained.
InfoMerge has substantial acceleration in both prefill and generation.
Under 15\% retention, InfoMerge achieves substantial acceleration, with a $4.24\times$ speedup in prefill and a $3.95\times$ speedup in generation while
maintaining 98.8\% of the model performance on LLaVA-OneVision~\cite{li2024llava-ov}.
This confirms that InfoMerge effectively improves inference efficiency under high compression rates.


\begin{table}[t]
\centering
\small
\setlength{\tabcolsep}{5.5pt}
\caption{
Ablation of temporal redundancy estimation on VideoMME.
``Adjacent Diff.'' follows DyCoke and computes first-order local feature differences within a temporal window of size 4, while TFD uses our second-order temporal fingerprint difference.
}
\label{tab:tfd_ablation}
\begin{tabular}{lccc}
\toprule
\textbf{Method} & \textbf{$r{=}5\%$} & \textbf{$r{=}10\%$} & \textbf{$r{=}15\%$} \\
\midrule
FastVID & 53.6 & 56.8 & 57.8 \\
+ Adjacent Diff. + CABA & \underline{54.4} & \underline{57.5} & \textbf{58.2} \\
+ TFD + CABA & \textbf{54.6} & \textbf{57.7} & \textbf{58.2} \\
\bottomrule
\end{tabular}
\end{table}

\begin{table}[t]
\centering
\small
\setlength{\tabcolsep}{6pt}
\caption{Ablation on the two-factor dynamic budget allocation on VideoMME. 
$\alpha$ and $\beta$ denote the weights of segment uniqueness and representational richness, respectively. Retain Ratio$=10\%$}
\begin{tabular}{c c c}
\toprule
$\alpha$ & $\beta$ & VideoMME Acc. \\
\midrule
1.00 & 0.00 & 57.6 \\
0.90 & 0.10 & \textbf{57.7} \\
0.80 & 0.20 & 57.4 \\
0.70 & 0.30 & 57.2 \\
0.60 & 0.40 & 57.0 \\
0.50 & 0.50 & 57.1 \\
0.30 & 0.70 & 57.4 \\
0.00 & 1.00 & 56.9 \\
\bottomrule
\end{tabular}
\label{tab:two_factor_ablation}
\vspace{-3mm}
\end{table}

\subsection{Ablation Study}
\label{sec:ablation}

\paragraph{Effect of Temporal Fingerprint Difference.}

We first evaluate the effectiveness of TFD for temporal redundancy estimation.
As shown in Table~\ref{tab:component_ablation}, it confirms the effectiveness of TFD as the basic redundancy prior: using TFD alone already improves over the original FastVID across all retention ratios.
As shown in Table~\ref{tab:tfd_ablation}, 
we further compare TFD with a first-order local difference baseline following DyCoke~\cite{tao2025dycoke}, which computes feature differences within a temporal window of size 4.
Compared with this first-order window-based variant, TFD achieves better performance at stricter retention ratios, indicating that second-order temporal fingerprints are more effective for identifying temporally stable redundant regions when the token budget is limited.

\paragraph{Effect of Content-Aware Budget Allocation.}
We then study the contribution of the proposed dynamic budget allocation strategy.
We first explore the fusion weights between segment uniqueness and representational richness to determine the best allocation configuration.
Specifically, we fix the overall visual token retention ratio to 10\% and vary the fusion weights of the two factors.
As shown in Table~\ref{tab:two_factor_ablation}, segment uniqueness is the stronger single factor, while representational richness provides complementary structural information.
Their combination achieves the best performance, suggesting that segment-level budget allocation benefits from jointly considering segment uniqueness and representational richness.
We further examine the contribution of each factor in Table~\ref{tab:component_ablation}.
Adding segment uniqueness to TFD improves the performance at 5\% and 10\% retention ratios, showing that segment-level uniqueness is useful for budget reallocation.
Adding representational richness provides complementary gains, especially at 15\% retention, suggesting that representational richness helps capture the internal visual richness of each segment.
The full model achieves the best results at 5\% and 10\% and remains competitive at 15\%, demonstrating the overall effectiveness of the two-factor dynamic budget allocation.

\begin{figure}[t]
    \centering
    \includegraphics[
         width=\linewidth,
         trim=60 10 0 20,
         clip
     ]{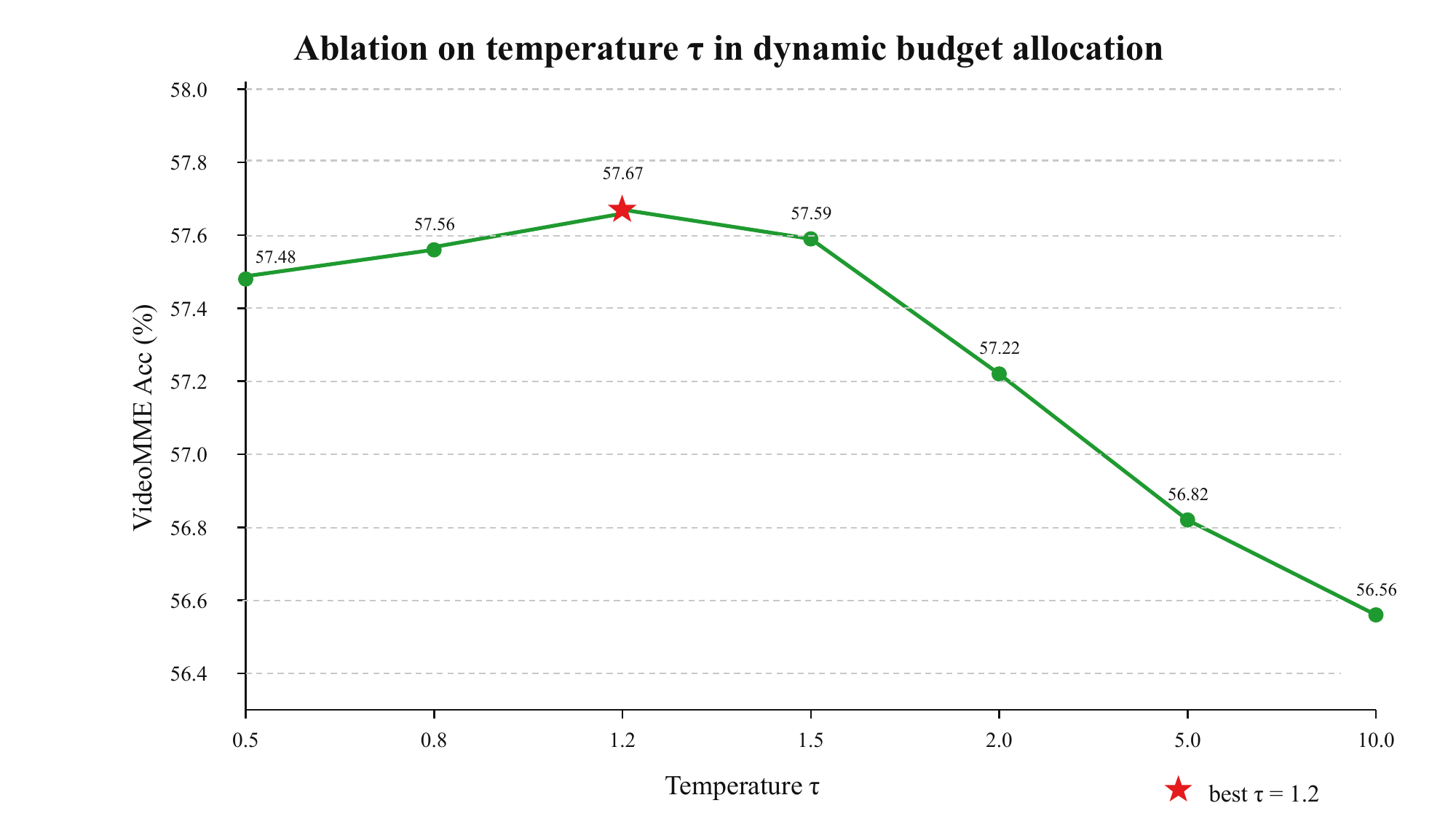}
    \caption{
    Ablation on the temperature parameter $\tau$ in dynamic budget allocation.
    Moderate temperature leads to the best trade-off between smooth and aggressive budget redistribution. Retain Ratio $=10\%$.
    }
    \label{fig:temp}
\end{figure}

\begin{figure}[t]
    \centering
    \includegraphics[width=\linewidth,trim=60 40 0 30,
         clip]{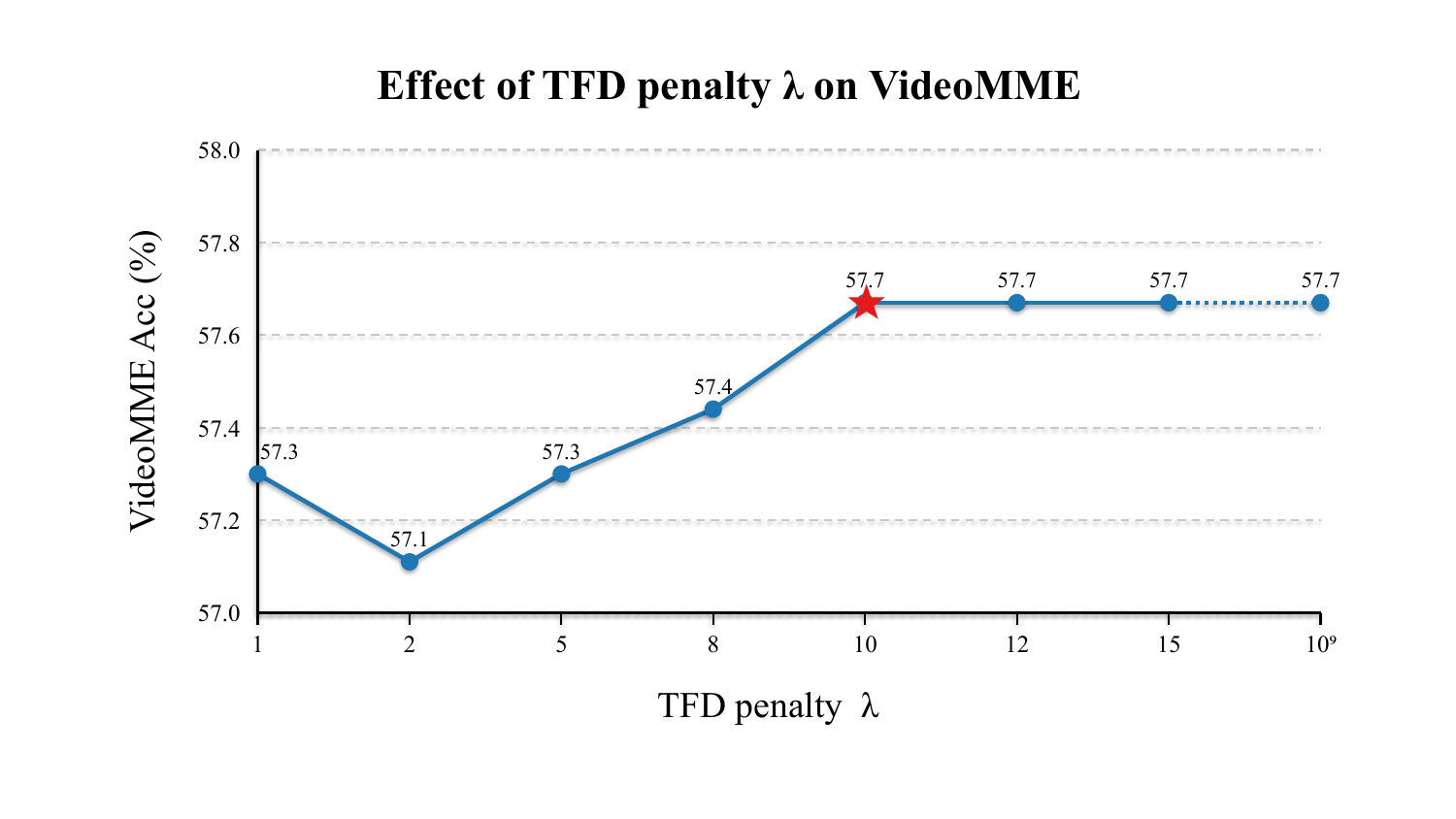}
    \caption{
    Ablation on the redundancy suppression strength $\lambda$ on VideoMME.
    Stronger suppression reduces repeated preservation of static tokens and improves token utilization. Retain Ratio $=10\%$.
    }
    \label{fig:pen}
\end{figure}

\paragraph{Sensitivity to Temperature.}

We further conduct an ablation study on the temperature parameter $\tau$ in dynamic budget allocation.
The temperature controls the sensitivity of the sigmoid modulation function to the segment score.
A smaller value makes the budget distribution smoother, while a larger value concentrates the token budget on a few high-scoring segments.
As shown in Figure~\ref{fig:temp}, a moderate temperature achieves the best performance, indicating that overly smooth or overly aggressive redistribution can both be suboptimal.

\paragraph{Effect of Redundancy Suppression Strength.}

Figure~\ref{fig:pen} analyzes the effect of the redundancy suppression strength $\lambda$.
A larger $\lambda$ more strongly penalizes previously selected static positions.
The results show that strong suppression works well, suggesting that repeatedly selected static regions contribute limited additional information and should be explicitly discouraged during token selection.



\section{Conclusion}
\label{sec:conclusion}
In this paper, we propose {InfoMerge}, a novel training-free token compression method for efficient Video LLM inference. Our method is built upon two key components: (1) a {Temporal Fingerprint Difference (TFD)} module, which performs segment-level second-order redundancy identification; and (2) a {Content-Aware Budget Allocation} strategy, which adaptively distributes token budgets according to both segment uniqueness and representational richness across video segments.  Extensive experiments across multiple Video-LLM architectures and benchmarks demonstrate the effectiveness and strong generalization ability of our method. 
Notably, InfoMerge exhibits more pronounced advantages under aggressive compression, achieving favorable efficiency--performance trade-offs.

\section{Limitations}

Although InfoMerge achieves a favorable efficiency--performance trade-off for Video LLM inference, several limitations remain.  
The proposed Temporal Fingerprint Difference (TFD) module relies on segment-level temporal statistics.
For regions with persistent jitter, the estimated temporal fingerprints may become less reliable, which can reduce the accuracy of identifying low-information regions such as static backgrounds.
Besides, although InfoMerge generalizes well across several Video LLM architectures, our experiments mainly focus on representative open-source models and benchmarks. 
Its effectiveness on larger-scale multimodal systems or highly domain-specific video distributions remains to be further explored. 
While the proposed method significantly reduces inference FLOPs and memory usage, the additional computation introduced by redundancy estimation and spectral analysis may still incur non-negligible overhead under extremely resource-constrained environments. Finally, during the preparation of this manuscript, we leveraged AI assistants (Gemini and ChatGPT) for text polishing and grammatical corrections.



\bibliography{latex/bib/custom}

\newpage
\clearpage

\appendix

\section{More Implementation Details}
\subsection{Additional Experimental Settings}
\label{app:exp_settings}

For latency measurement, we use batch size 1 and report the average latency after warm-up runs with CUDA synchronization.
Latency is measured after visual features are extracted, excluding video decoding, frame sampling, vision-tower encoding, and projector computation.
Therefore, the reported speedups reflect LLM-side acceleration rather than full end-to-end video-QA latency.

\subsection{FLOPs Calculation}
\label{app:flops}

We report MAC-style FLOPs, where one multiply-accumulate operation is counted as one operation.
Therefore, all FLOPs numbers are not multiplied by the conventional $2\times$ MAC-to-FLOP conversion factor.
We count only the prefill computation of the Qwen2 language backbone and exclude the vision encoder, projector, embedding lookup, RoPE, RMSNorm, softmax, and the LM head.

For a Qwen2 decoder layer with sequence length $N$, hidden size $d$, intermediate size $d_{\mathrm{ff}}$, number of attention heads $H$, and number of key-value heads $H_{\mathrm{kv}}$, the MAC-style FLOPs are computed as:
\begin{equation}
\begin{aligned}
\mathrm{FLOPs}_{\mathrm{layer}}(N)
&= N d^2
\left(
2 + 2\frac{H_{\mathrm{kv}}}{H}
\right) \\
&\quad + 2N^2d
+ 3Ndd_{\mathrm{ff}} .
\end{aligned}
\end{equation}

The first term counts the Q/K/V/O projections under grouped-query attention, the second term counts the two attention matrix multiplications, i.e., $QK^\top$ and $\mathrm{Attn}V$, and the last term counts the SwiGLU MLP projections.

For an $L$-layer Qwen2 backbone, the total prefill FLOPs are:
\begin{equation}
\mathrm{FLOPs}(N)
=
L \cdot \mathrm{FLOPs}_{\mathrm{layer}}(N).
\end{equation}
For Qwen2-7B, we use:
$
L=28,\quad
d=3584,\quad
d_{\mathrm{ff}}=18944,
\quad
H=28,\quad
H_{\mathrm{kv}}=4.
$
For outer-LLM token compression methods such as FastVID and InfoMerge, visual tokens are compressed before being fed into the Qwen2 backbone.
Therefore, all decoder layers use the reduced sequence length $N_r$:
\begin{equation}
\mathrm{FLOPs}_{\mathrm{outer}}
=
L \cdot \mathrm{FLOPs}_{\mathrm{layer}}(N_r).
\end{equation}
In contrast, for FastV, token pruning is applied after the first $K$ full-token decoder layers.
Following FastV~\cite{chen2024image}, we set $K=2$ and compute:
\begin{equation}
\mathrm{FLOPs}_{\mathrm{FastV}}
=
K \cdot \mathrm{FLOPs}_{\mathrm{layer}}(N_0)
+\\
(L-K) \cdot \mathrm{FLOPs}_{\mathrm{layer}}(N_r),
\end{equation}

where $N_0$ is the original sequence length and $N_r$ is the sequence length after pruning.
In our visual-token-only FLOPs analysis, the original sequence length is computed as:
\begin{equation}
N_0 = F \times P,
\end{equation}
where $F$ denotes the number of sampled frames and $P$ denotes the number of visual tokens per frame.
The reduced sequence length $N_r$ is determined by the corresponding visual-token retention ratio.
Unless otherwise stated, text tokens are excluded from the FLOPs calculation.

\subsection{Source of Baseline Results}
Unless otherwise stated, all results reported in our tables are evaluated locally under the same evaluation protocol.
For Table~\ref{tab:llava_video_sota}, the accuracy results of FastV, VisionZip, and MMG-Vid on LLaVA-Video-7B are taken from MMG-Vid~\cite{ma2026mmg}.
The FastVID and InfoMerge results in Table~\ref{tab:llava_video_sota} are evaluated locally under the same 64-frame setting.
For all other tables, we locally evaluate all compared methods except MMG-Vid, whose results are taken from its original paper, since MMG-Vid is not publicly available at the time of our experiments.

\subsection{Details of Token Budget Decomposition}
\label{app:budget_decomposition}

Given an overall visual-token retention ratio $r$, the total retained token budget for a video with $L$ sampled frames and $N$ visual tokens per frame is
\begin{equation}
B_{\mathrm{total}} = rLN .
\end{equation}
Following FastVID~\cite{shen2026fastvid}, we use a hyperparameter $d\in[0,1]$ to split the retained tokens into two branches:
\begin{equation}
B_{\mathrm{DTM}} = dB_{\mathrm{total}},
\quad
B_{\mathrm{ATS}} = (1-d)B_{\mathrm{total}} .
\end{equation}
Here, $B_{\mathrm{ATS}}$ is used for preserving fine-grained salient tokens selected by ATS, while $B_{\mathrm{DTM}}$ is used for maintaining segment-level contextual information through DTM.

For ATS, the salient-token budget is uniformly assigned to each frame:
\begin{equation}
T_s =
\left\lfloor
\frac{B_{\mathrm{ATS}}}{L}
\right\rfloor
=
\left\lfloor
(1-d)rN
\right\rfloor ,
\end{equation}
where $T_s$ denotes the number of salient tokens selected by ATS in each frame.

For DTM, the context-token budget is distributed across temporal segments according to the CABA described in Section~\ref{sec:method:budget}.
Given the segment allocation weight $w_k$, the context-token budget for the $k$-th segment is computed as
\begin{equation}
B_k =
\max
\left(
1,
\operatorname{round}
\left(
\frac{B_{\mathrm{DTM}} w_k}
{\sum_{j=1}^{K} w_j}
\right)
\right).
\end{equation}
In this way, the overall retention ratio $r$ determines the total compression strength, while $d$ controls the trade-off between ATS-based salient-token preservation and DTM-based context-token aggregation.

\subsection{Details of Dynamic Temporal Segmentation}
\label{app:dyseg}

We follow FastVID~\cite{shen2026fastvid} and use Dynamic Temporal Segmentation (DySeg) to partition sampled video frames into temporally ordered segments.
Given a video with $L$ sampled frames, let $\mathbf{g}_l$ denote the frame-level global feature of the $l$-th frame.
DySeg computes the transition similarity between adjacent frames as:
\begin{equation}
t_l = \cos(\mathbf{g}_l, \mathbf{g}_{l+1}), 
\quad l = 1,\dots,L-1 .
\end{equation}
A smaller $t_s$ indicates a stronger semantic transition between two adjacent frames and is therefore more likely to be selected as a segment boundary.

Let $\mathcal{T}=\{t_l\}_{l=1}^{L-1}$ denote all transition similarities.
DySeg combines a minimum segment number constraint and a similarity threshold to determine the final boundary set:
\begin{equation}
\mathcal{S}_1 = \operatorname{argmin}_{c-1} \mathcal{T},
\quad
\\
\mathcal{S}_2 = \{l \mid t_l < \tau_{\mathrm{seg}}\},
\quad
\mathcal{S} = \mathcal{S}_1 \cup \mathcal{S}_2 .
\end{equation}

Here, $c$ denotes the minimum number of segments, and $\tau_{\mathrm{seg}}$ is the transition-similarity threshold.
$\mathcal{S}_1$ ensures that each video is partitioned into at least $c$ segments, while $\mathcal{S}_2$ further introduces boundaries at obvious semantic transitions.
The resulting boundary set $\mathcal{S}$ divides the sampled frames into $K$ temporally ordered segments $\{V_k\}_{k=1}^{K}$.
All subsequent TFD-based redundancy estimation, information-density-aware budget allocation, and spatio-temporal token compression are performed independently within these segments.

\subsection{Details of Content-Aware Budget Allocation}
\label{app:caba}

\paragraph{Spectral entropy as representational richness.}
In CABA, we use normalized spectral entropy to estimate the internal representational richness of each segment.
Given the flattened segment feature matrix $\mathbf{X}_k \in \mathbb{R}^{|\mathcal{V}_k| N \times D}$, let $\{\sigma_j\}_{j=1}^{R}$ denote its singular values, where $R=\min(|\mathcal{V}_k|N,D)$.
We define the normalized spectral energy distribution as
\begin{equation}
\pi_j
=
\frac{\sigma_j^2}
{\sum_{\ell=1}^{R}\sigma_\ell^2}.
\end{equation}
We use squared singular values because they correspond to the spectral energy of the feature matrix.
According to the Frobenius norm decomposition,
\begin{equation}
\|\mathbf{X}_k\|_F^2
=
\sum_{j=1}^{R}
\sigma_j^2,
\end{equation}
where $\sigma_j^2$ measures the amount of feature energy captured by the $j$-th singular direction.
Therefore, $\{\pi_j\}_{j=1}^{R}$ can be interpreted as an energy distribution over different principal directions.

Based on this distribution, we compute the normalized spectral entropy:
\begin{equation}
e_k
=
\frac{
-\sum_{j=1}^{R}
\pi_j \log \pi_j
}{
\log R
}.
\end{equation}
The normalization term $\log R$ ensures that $e_k$ lies in a comparable range across segments with different effective matrix sizes.
A small $e_k$ indicates that most energy is concentrated in a few dominant directions, suggesting a lower-dimensional and more redundant representation.
In contrast, a large $e_k$ indicates that the energy is distributed across more principal directions, suggesting richer and more complex visual semantics.

\begin{figure*}[t]
    \centering
    \includegraphics[width=\linewidth]{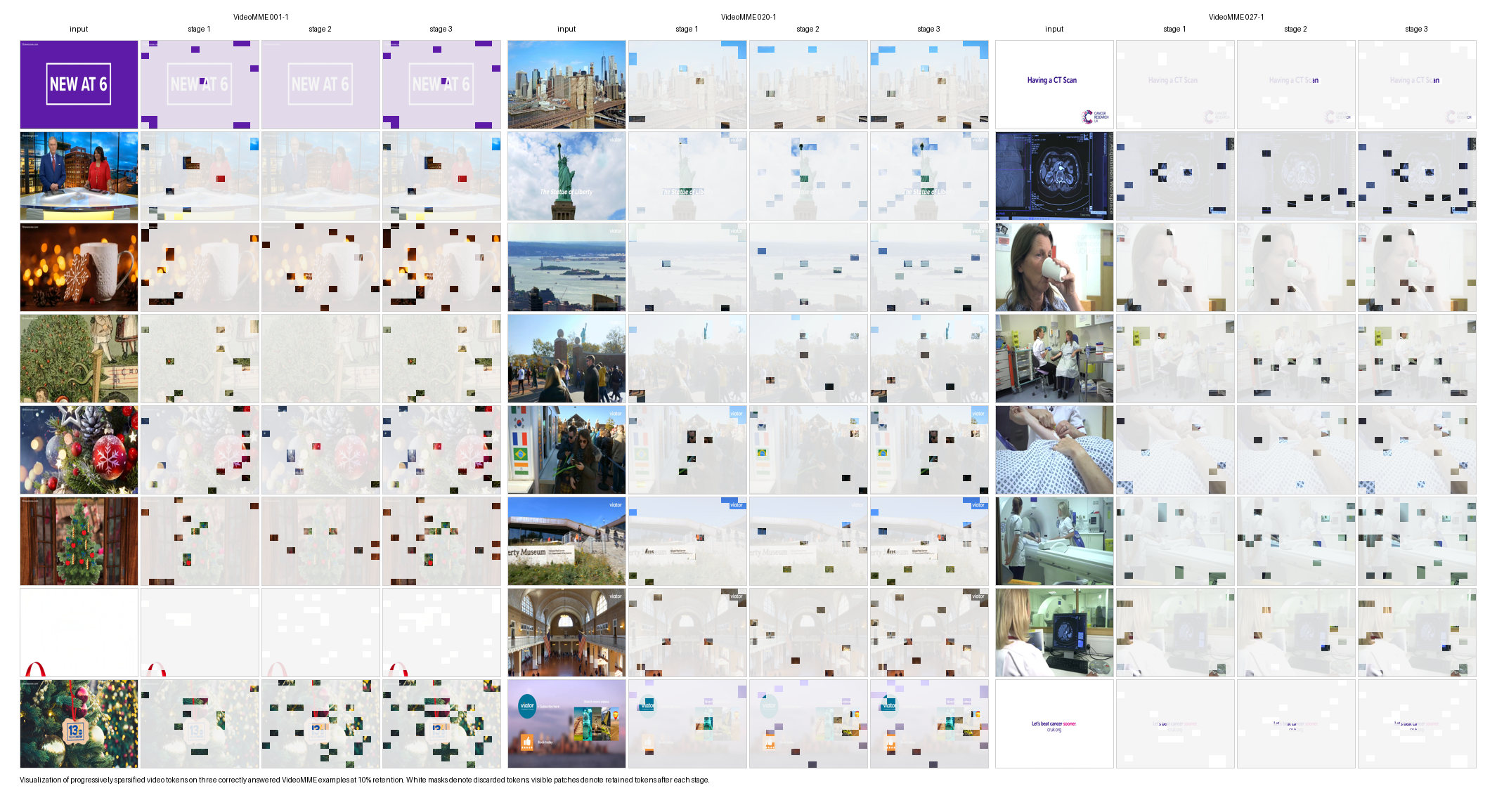}
\caption{Visualization of token selection in the proposed InfoMerge.Segments with lower information content receive fewer context tokens, while informative segments receive larger budgets.
}
    \label{fig:tokenSelect}
\end{figure*}

\paragraph{Relation to segment-level budget allocation.}
Segment uniqueness and representational richness capture complementary aspects of segment information.
The uniqueness score $u_k$ measures how much a segment deviates from the global video representation, while $e_k$ measures the diversity of internal feature directions within the segment.
Since the two factors have different numeric ranges, we apply z-score normalization before fusion:
\begin{equation}
m_k
=
\alpha z(u_k)
+
\beta z(e_k).
\end{equation}
The sigmoid function with temperature $\tau$ controls the sharpness of redistribution:
\begin{equation}
w_k
=
\sigma(\tau m_k)\cdot |\mathcal{V}_k| .
\end{equation}
Figure~\ref{fig:tokenSelect} provides a qualitative visualization of the proposed content-aware budget allocation.
We observe that segments with lower information content, such as visually repetitive or less event-relevant clips, are assigned fewer context tokens.
In contrast, segments containing key visual evidence or more diverse visual content receive larger token budgets.
This suggests that InfoMerge redistributes the limited token budget toward more informative temporal segments.

\begin{figure*}[t]
    \centering
    \includegraphics[
        width=0.95\textwidth,
        trim=60 0 10 0,
        clip
    ]{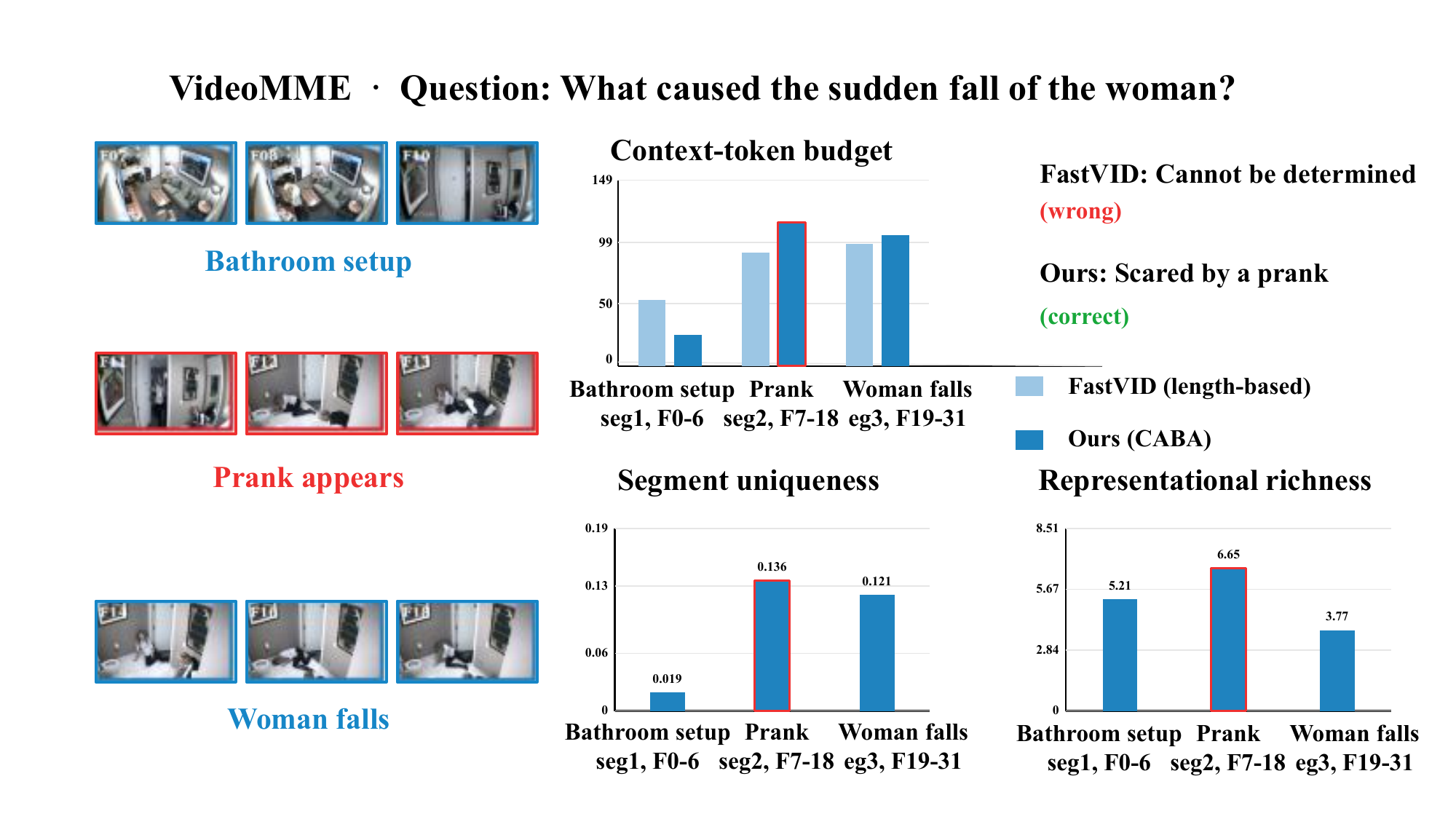}
    \caption{
    Case study of evidence-aware budget allocation on VideoMME.
The question asks what caused the sudden fall of the woman, requiring the model to identify the causal event before the fall.
We show three temporal segments corresponding to the bathroom setup, the prank appearance, and the woman's fall.
Compared with FastVID's length-based allocation, InfoMerge assigns more context-token budget to the evidence-critical prank segment, which also exhibits higher segment uniqueness and representational richness.
This helps preserve the key visual evidence for causal reasoning.
As a result, InfoMerge correctly predicts ``Scared by a prank'', while FastVID fails to determine the cause.
    }
    \label{fig:evidence_budget_case}
\end{figure*}

\subsection{Details of Token Compression}
\paragraph{Static-token replacement in ATS.}
\label{app:ats}
Since many Video-LLMs employ SigLIP as the visual encoder, the corresponding [CLS] attention is unavailable during inference. Following FastVID~\cite{shen2026fastvid}, we re-attach the pretrained SigLIP head to obtain lightweight saliency estimation without modifying the original Video-LLM pipeline.
When a selected ATS token position is identified as static by TFD, we reuse a segment-level merged representation for this spatial position to avoid repeatedly preserving near-duplicate static tokens.
Specifically, for position $p\in \mathcal{T}^{\mathrm{ATS},(k)}_l\cap \mathcal{S}^{(k)}$, we compute a segment-level static representation by aggregating tokens at the same spatial position across the $k$-th segment:
\begin{equation}
\mathbf{m}^{(k)}_{p}
=
\frac{1}{|\mathcal{V}_k| }
\sum_{r=1}^{|\mathcal{V}_k| }
\mathbf{H}^{(k)}_{r,p}.
\end{equation}
Then, for the current frame and all subsequent frames in the same segment, the token at position $p$ is replaced by this merged representation:
\begin{equation}
\mathbf{H}^{(k)}_{r,p}
\leftarrow
\mathbf{m}^{(k)}_{p},
\quad
r=l,\dots,|\mathcal{V}_k|.
\end{equation}
This operation reduces repeated preservation of visually similar static tokens while keeping a compact representation of the corresponding static region.
This replacement is a feature-level substitution and does not introduce an additional token-pruning step or change the number of retained tokens.

\paragraph{Density-peak anchor selection in DTM.}
Within each anchor frame, DTM follows density-peak clustering to select representative context anchors.
For token $t_i$, its local density is computed as
\begin{equation}
\rho_i
=
\exp
\left(
-
\frac{1}{K_{\mathrm{nn}}}
\sum_{t_j\in\mathrm{kNN}(t_i)}
d(t_i,t_j)^2
\right),
\end{equation}
where $d(\cdot,\cdot)$ denotes the token distance and $\mathrm{kNN}(t_i)$ denotes the $K_{\mathrm{nn}}$ nearest neighbors of $t_i$.
The distance to higher-density tokens is defined as
\begin{equation}
\delta_i=
\begin{cases}
\min_{j:\rho_j>\rho_i}
d(t_i,t_j),
&
\exists j,\rho_j>\rho_i,
\\
\max_j d(t_i,t_j),
&
\text{otherwise}.
\end{cases}
\end{equation}
The final density-peak score is
\begin{equation}
s_i=\rho_i\delta_i.
\end{equation}
Tokens with higher density-peak scores are selected as context anchors unless suppressed by the TFD-guided hard mask described in Section~\ref{sec:method:compression}.

\paragraph{Anchor-centric aggregation.}
After context anchors are selected, each non-anchor token is assigned to its nearest anchor token center.
For anchor token $a$ with assigned token set $\{b_1,\dots,b_n\}$, the updated anchor representation is computed as
\begin{equation}
a^\star
=
\mu a
+
(1-\mu)
\frac{1}{n}
\sum_{j=1}^{n}b_j,
\end{equation}
where $\mu$ balances the original anchor feature and the aggregated neighboring information.
The updated anchors are used as contextual tokens and concatenated with the salient tokens selected by ATS.

\section{Qualitative Analysis}
\label{sec:qualitative}
To better understand the effect of content-aware budget allocation, we provide a qualitative case study in Figure~\ref{fig:evidence_budget_case}.
The question asks what caused the sudden fall of the woman, which requires the model to identify the causal event before the fall.
FastVID allocates context-token budgets mainly according to segment length and may under-preserve short but evidence-critical segments.
In contrast, InfoMerge assigns more context tokens to the prank segment according to its segment-level information content, thereby preserving key visual evidence for causal reasoning.
As a result, InfoMerge correctly predicts ``Scared by a prank'', while FastVID fails to determine the cause.
This example demonstrates that InfoMerge can redistribute the limited token budget toward informative temporal segments instead of relying only on segment length.



\begin{table}[t]
\centering
\caption{Hyperparameter settings under different visual-token retention ratios.}
\label{tab:hy}
\begin{tabular}{c c c c c c}
\toprule
$r$ & $\tau$ & $\rho$ &
$\lambda$ & $\alpha$ & $\beta$ \\
\midrule
0.05 & 2.0 & 0.05 & 12.0 & 0.9 & 0.1 \\
0.10 & 1.2 & 0.10 & 12.0 & 0.9 & 0.1 \\
0.15 & 1.2 & 0.09 & 12.0 & 0.9 & 0.1 \\
\bottomrule
\end{tabular}
\end{table}
\section{Hyperparameter Settings}
\label{app:hy}
We use different hyperparameter settings for different retention ratios to balance salient-token preservation and contextual-token aggregation under varying compression strengths.
The detailed settings are summarized in Table~\ref{tab:hy}.
Unless otherwise specified, all methods are evaluated under the same frame sampling strategy and evaluation protocol.

\end{document}